%% file: main.tex
\documentclass[10pt,twocolumn,letterpaper]{article}

\usepackage[pagenumbers]{cvpr} 


\definecolor{cvprblue}{rgb}{0.21,0.49,0.74}
\usepackage[pagebackref,breaklinks,colorlinks,allcolors=cvprblue]{hyperref}
\usepackage{multirow}
\usepackage{algorithm}
\usepackage{algpseudocode}

\newcommand{\ModuleName}{\textbf{AnySynth}}

\title{AnySynth: Harnessing the Power of Image Synthetic Data Generation for Generalized Vision-Language Tasks}

\author{
    You Li \quad
    Fan Ma \quad
    Yi Yang$^\dag$ \\
    ReLER, CCAI, Zhejiang University, Zhejiang, China\\
    {\small $^\dag$ Corresponding author} \\
    {\tt\small {\{uli2000, mafan, yangyics\}}@zju.edu.cn} \\
}

\begin{document}

\maketitle
\input{sec/0_abstract}    
\input{sec/1_intro}
\input{sec/2_related_work}
\input{sec/3_method}

\input{sec/4_experiment}

\input{sec/5_conclusion}
{
    \small
    \bibliographystyle{ieeenat_fullname}
    \bibliography{main}
}

\input{sec/X_suppl}

\end{document}

%% file: sec/0_abstract.tex
\begin{abstract}

Diffusion models have recently been employed to generate high-quality images, reducing the need for manual data collection and improving model generalization in tasks such as object detection, instance segmentation, and image perception.
However, the synthetic framework is usually designed with meticulous human effort for each task due to various requirements on image layout, content, and annotation formats,  restricting the application of synthetic data on more general scenarios.
In this paper, we propose \ModuleName, a unified framework integrating adaptable, comprehensive, and highly controllable components capable of generating an arbitrary type of synthetic data given diverse requirements.
Specifically, the Task-Specific Layout Generation Module is first introduced to produce reasonable layouts for different tasks by leveraging the generation ability of large language models and layout priors of real-world images.
A Uni-Controlled Image Generation Module is then developed to create high-quality synthetic images that are controllable and based on the generated layouts. In addition, user specific reference images, and style images can be incorporated into the generation to task requirements. 
 Finally, the Task-Oriented Annotation Module offers precise and detailed annotations for the generated images across different tasks.
We have validated our framework's performance across various tasks, including Few-shot Object Detection, Cross-domain Object Detection, Zero-shot Composed Image Retrieval, and Multi-modal Image Perception and Grounding. 
The specific data synthesized by our framework significantly improves model performance in these tasks, 
demonstrating the generality and effectiveness of our framework.
\end{abstract}

%% file: sec/1_intro.tex
\section{Introduction}
\label{sec:intro}

\begin{figure}[tb]
    \setlength{\abovecaptionskip}{0.cm}
    \setlength{\belowcaptionskip}{0.cm}
    \begin{center}
        \includegraphics[width=0.48\textwidth]{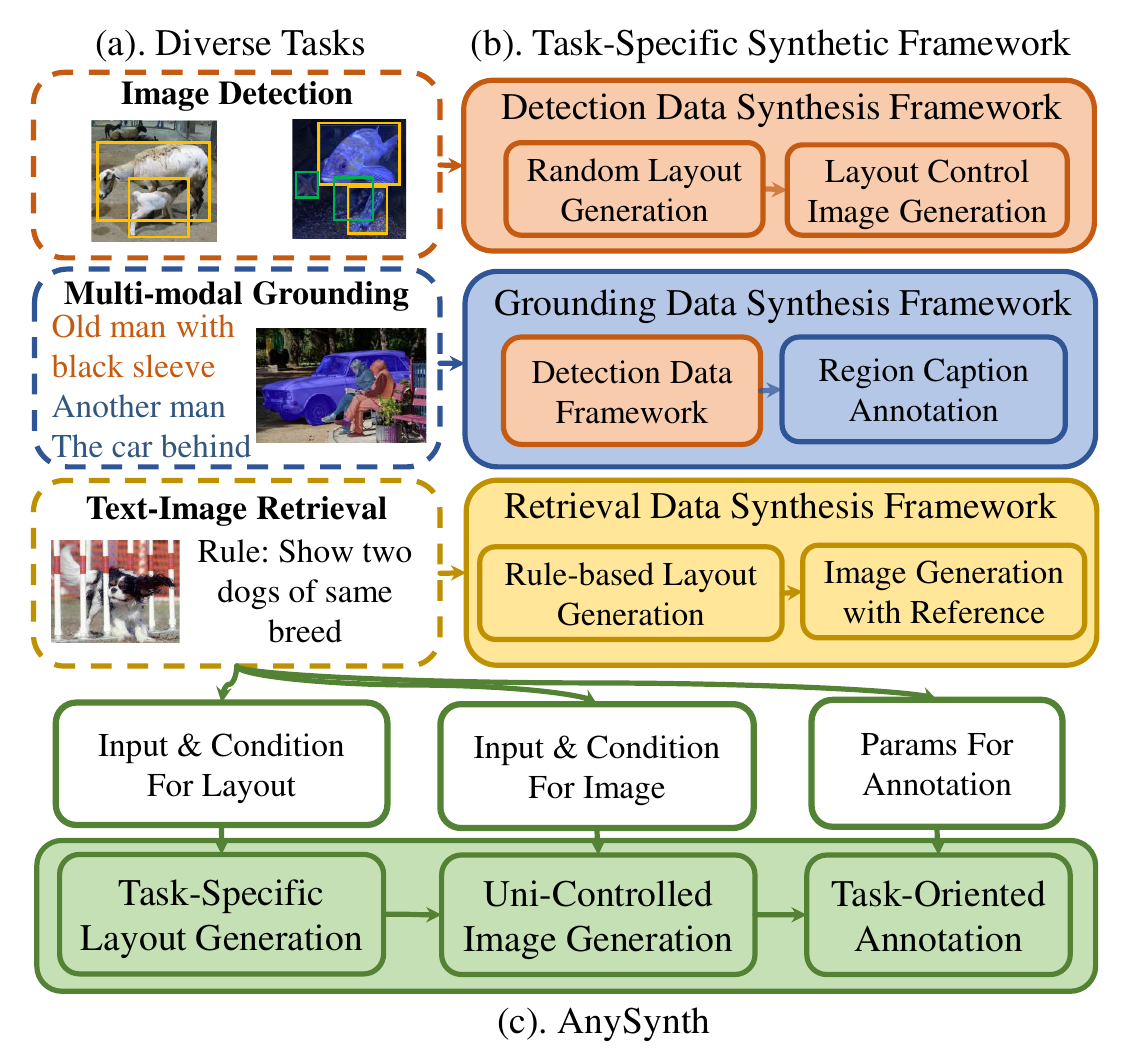}
    \end{center}
   \caption{\textbf{The Differences of AnySynth from Other Frameworks.} In (a), we showcase several typical tasks with diverse requirements, in (b), we show the common Synthetic data collection frameworks, which need specific desgin on different tasks. In (c), we show our AnySynth Framework, handle diverse tasks in one unified framework, enhance the generality of synthetic data.
} 
    \label{fig:show}
  \end{figure}

Recent significant advancements have been made in fields such as computer vision~\cite{DALL-E2,CLIP,pydiff,stablediffusion,caphuman,3dis,li2024imagine} and multi-modal understanding~\cite{llava,llavanext,improvedllava,glamm}, largely driven by the well-annotated real-world image data~\cite{coco,voc,roboflow}. 
\textit{\textbf{The traditional real data collection framework}} typically involves collecting task-specific real images from the internet, followed by annotation and filtering, either manually or with automation. Although human intervention ensures the data quality, the process remains \textbf{labor-intensive} and \textbf{time-consuming}, which limits the scalability of datasets. While introducing automated machine annotation reduces some costs, it remains constrained by the volume and domain of the collected real data.

Diffusion model-based image generation methods~\cite{ddpm,stablediffusion,glide,guideddiffusion,Ho2022ClassifierFreeDG} can produce high-quality images. The development of controllable generation techniques~\cite{boxdiff,gligen,instancediffusion,migc,migc++} has further empowered users to steer the generated content. Therefore, many frameworks~\cite{acp, odgen,gen2det,aug1,aug2} attempt to synthesize training data for downstream tasks. However, As shown in Fig.\ref{fig:show} (a), \textbf{with the downstream tasks change, these frameworks need manual adjustments or redesign}. For example, users may need to adjust certain modules within the existing framework to generate the required annotations. When faced with special training data pairs, more control conditions may be introduced, which could even require a complete redesign of framework. This limits the applicability and impact of data synthesis methods. Consequently, \textit{developing \textbf{Unified}} framework that can synthesize \textbf{high-quality data suitable for various tasks} becomes a crucial research focus.

We believe that the difference in requirements for data mainly varies in \textit{image layout, content, and annotation formats}, which can be addressed through \textit{adaptable, comprehensive, and highly controllable components} with \textbf{the relevant inputs and control conditions}. We thus propose \ModuleName, a Unified synthetic data collection framework that generates data suitable for various downstream tasks. As shown in Fig.\ref{fig:show} (b), \ModuleName\ adopts a Layout-Image-Annotation structure, progressively synthesizes data with the inputs and control conditions of each stage.
The first core step is to generate a realistic scene and layout that are both reasonable and align with specific requirements. Therefore, we propose the \textit{\textbf{Task-Specific Layout Generation Module}}, which utilizes LLM to understand various conditions and inputs and generate appropriate layouts. The second step involves synthesizing the image based on various conditions, such as layout, text, reference images, and style images. We propose the \textit{\textbf{Uni-Controlled Image Generation Module}}, introducing multiple conditions like reference and style images into the controllable generation methods MIGC, improve the quality of images through position and quality based filtering. Furthermore, \textit{\textbf{The Task-Oriented Annotation Module}} offers task-specific annotations required for different tasks.

We demonstrated the effectiveness of synthetic data in \textit{\textbf{data-scarce}} scenarios through a series of Few-shot tasks. In the Few-shot Object Detection task, under the 10-shot setting for VOC's Split1, Split2, and Split3, we raised the AP for novel classes from 68.2/53.5/60.9 to 69.1/55.0/63.8. On the COCO dataset, we further improved the AP for 10/30 shots from 15.8/19.6 to 18.8/21.0. Additionally, we selected the Zero-Shot Composed Image Retrieval task to validate the effectiveness of synthetic data in \textit{\textbf{various compositional forms}}, in which we generated synthetic images based on reference images and relative captions. On the CIRCO dataset, mAP for k=5 rose from 23.35 to 24.57, while recall for k=1 on the CIRR dataset improved from 26.53 to 27.33. Expanding to \textit{\textbf{cross-domain}} tasks, we increased YOLOv5's AP50 on the Roboflow Aquarium data from 47.4 to 49.9, and on the People-in-Art dataset from 21.2 to 22.7. We also validated our framework on \textit{\textbf{multi-modal large models}}, enhancing the METEOR from 16.1 to 16.7 and CIDEr from 106.0 to 107.6 in Region Captioning on GaLMM. Finally, ablation experiments validated the performance of each module.

Our contributions are summarized as follows:

\begin{itemize}
  \item [1)] 
  We propose \ModuleName, a unified data synthesis framework that can generate image data suitable for various downstream tasks.
  \item [2)]
  We have designed a Task-Specific Layout Generation module, a Uni-Controlled Image Generation Module, and a Task-Oriented Annotation Module capable of synthesizing the required data progressively and controllably.
  \item [3)]
  We validated our framework on various downstream tasks. The improvements in each experiment demonstrate its effectiveness, while the diverse experimental settings further confirm its high generality.
\end{itemize}

%% file: sec/2_related_work.tex
\section{Related work}
\label{sec:related_work}

\subsection{Text-to-Image Generation}

Text-to-image models\cite{gan,attngan,DALL-E2,ddpm,stablediffusion,glide,eDiff-I,guideddiffusion,Ho2022ClassifierFreeDG,reimagen,MuseTG,imagen} have seen rapid development in recent years, among them, diffusion-based models\cite{ddpm,stablediffusion,glide,eDiff-I,guideddiffusion,Ho2022ClassifierFreeDG} dominate, capable of generating high-quality images. Diffusion models treat image generation as a stepwise denoising process starting from noise sampled from a normal distribution, using UNet\cite{unet} or Transformer\cite{attention} architectures to predict and denoising at each step, ultimately producing high-quality images. Stable Diffusion\cite{stablediffusion} is a widely used foundational image generator that performs denoising operations in the latent space of images, enabling the generation of relatively high-quality images at a lower cost.

Generators like Stable Diffusion often struggle with limited textual control and face notable issues such as attribute confusion and missing objects\cite{migc}. As a result, recent works have sought to integrate layout information for instance-controllable text-to-image generation\cite{gligen,layoutdiff,tflcg,instancediffusion,migc,migc++}. For example, GLIGEN\cite{gligen} adds gated self-attention layers to improve object position control, while Instance Diffusion\cite{instancediffusion} further refines GLIGEN's framework, enhancing control over object attributes. MIGC\cite{migc,migc++} employs a divide-and-conquer approach to decompose complex text generation into the generation of individual instances, adding extra EA and LA layers to improve positional control.  These works enhance the content controllability of generated images without sacrificing quality.

\subsection{Synthetic data augmentation solution}

Data is crucial in deep learning, particularly for computer vision and multi-modal reasoning, while manual collection and annotation can be excessively costly. Consequently, many efforts aim to leverage synthetic data to address data scarcity. Gen2Det\cite{gen2det} focuses on constructing datasets for object detection tasks with stable diffusion, and ODGEN\cite{odgen} trains cross-domain object detection tasks after fine-tuning on specific domain images. Additionally, the ACP framework\cite{acp} leverages large language models(LLM)\cite{gpt4} to generate layouts corresponding to the object categories provided by the user and generate images for object detection and image reasoning. However, these existing synthetic data collection frameworks are task-specific, which restricts their applicability and reduces their overall practicality and impact.

%% file: sec/3_method.tex
\section{Method}
\label{sec:method}

\begin{figure*}[tb]
    \setlength{\abovecaptionskip}{0.cm}
    \setlength{\belowcaptionskip}{0.cm}
    \begin{center}
        \includegraphics[width=1.0\textwidth]{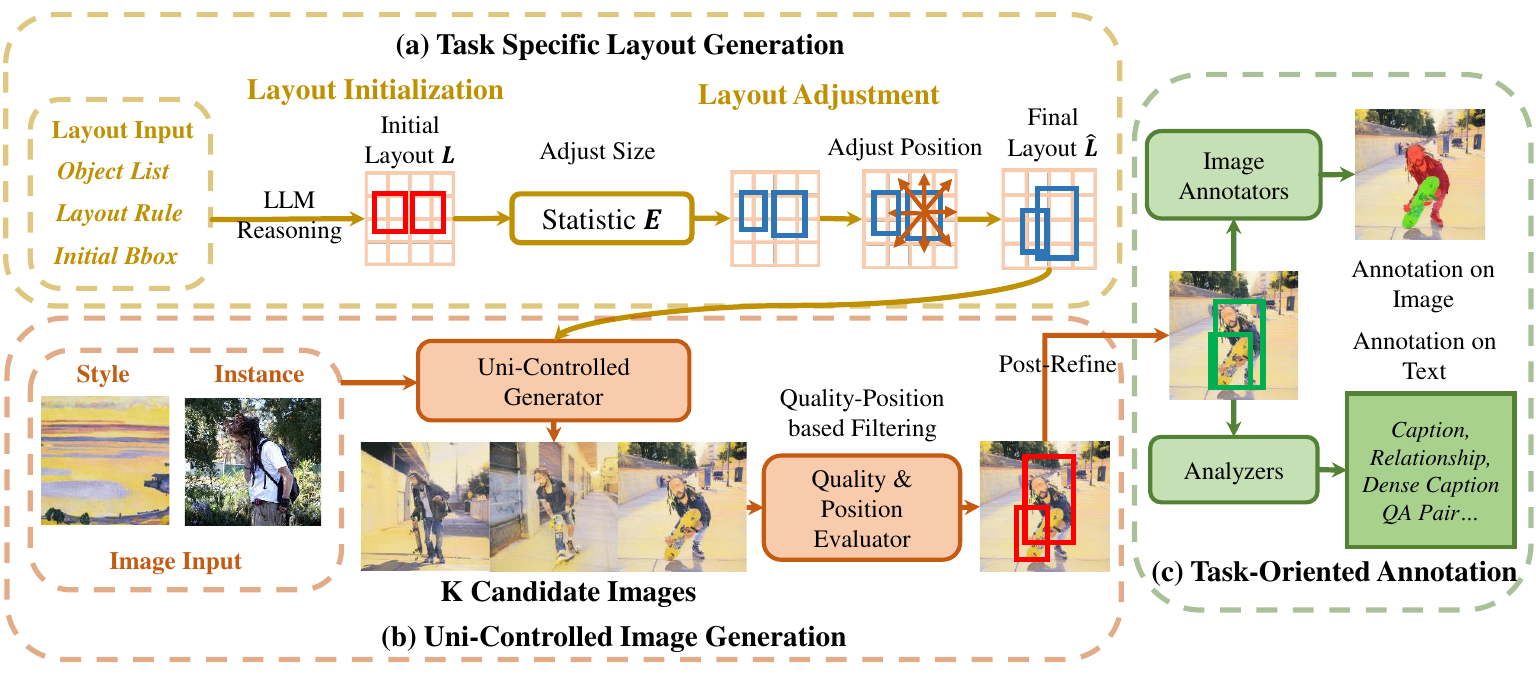}
    \end{center}
   \caption{\textbf{Overview of our \ModuleName}. Our \ModuleName\ consists of three modules. (a) Represents \textbf{the Task-Specific Layout Generation Module}, which parses various layout parameters using LLMs and combines them with dataset statistics to derive object layouts and basic scenes. (b) Represents \textbf{the Uni-Controlled Image Generation Module}, which achieves comprehensive high-quality image generation by integrating layout, instance reference and style reference image, and position quality filtering. (c) Represents \textbf{the Task-Oriented Annotation module}, which provides fine-grained annotations for downstream tasks. 
} 
  \vspace{-5pt}
    \label{fig:fig_2}
  \end{figure*}
  
Our \ModuleName\ is a Unified data synthesis framework, able to synthesize images for different tasks. To handle the different requirements in image layout, content, and annotations, we progressively synthesize images using the Layout-Image-Annotation structure, processing specific inputs and conditions at each step. We first provide the overview of our framework in section \ref{sec:input}. In section \ref{sec:layout}, we present the Task-Specific Layout Generation Module that combines LLMs and statistical data to generate a suitable layout. In section \ref{sec:generation}, we discuss how to generate high-quality image data with various conditions, and in section \ref{sec:region}, we introduce how to apply our framework to downstream tasks.

\subsection{Overview}
\label{sec:input}

To clarify the overall framework, inputs, and process, we first define the various input parameters in the generation process and systematically describe our entire synthesis framework in this section.

\noindent \textbf{Inputs and Conditions.} Compared to other synthetic data collection frameworks \cite{acp,gen2det,odgen} that only take object categories as input, our general synthetic framework allows for more comprehensive input as control. The main input information for our designed framework is as follows:

\begin{itemize}
  \item [1)] 
  \noindent \textbf{Object Categories $O$.} The object category parameters typically define which categories to sample from and the number of objects sampled.
  \item [2)]
  \noindent \textbf{Layout Rules $ R$.} Users can enhance their control over the generated layouts by providing specific layout instructions, such as replacing or adding certain objects within a particular layout.
  \item [3)]
  \noindent \textbf{Initial Bbox $ B $}: The layout of the target image may be adjusted based on an existing initial bboxes.
  \item [4)]
  \noindent \textbf{Style Image $I_s$.} Users can provide style images from the desired domain for generation, such as cartoons.
  \item [5)]
  \noindent \textbf{Instance Reference Image $I_i$.} Users can provide the corresponding visual reference image when generating a specific instance.
  \item [6)]
  \noindent \textbf{Annotation Formats $T$.} Different tasks require different annotations. For example, object detection needs bounding boxes, instance segmentation requires object masks, and multi-modal grounding demands region-based annotations. These inputs are primarily binary control signals.

\end{itemize}

    \noindent \textbf{Generation Process.} \ModuleName\ $\mathcal{F}$ follows the `Layout-Generation-Annotation' steps to synthesize data gradually. As shown in Fig.\ref{fig:fig_2} (a), \ModuleName\ first generates Layout $\hat{L}$ including scenes $S$ and instance layout $L_{inst}$ that conform to object categories $O$, Layout Rule $R$ and Initial Bbox $B$ with the Task-Specific Layout Generation Module $\mathcal{F}_l$:

\begin{equation}
    \hat{L} = (S, \hat{L}_{inst}) = \mathcal{F}_l(O, R, B). 
\end{equation}

Next, in Fig.\ref{fig:fig_2} (b), based on the scene and layout, the Uni-Controlled Generation Module $\mathcal{F}_g$ produces high-quality images $\mathcal{I}$ along with conditions including Layout $\hat{L}$, reference images $I_i$ and style images $I_s$:

\begin{equation}
    \mathcal{I}  = \mathcal{F}_g(I_s, I_i, \hat{L}).
\end{equation}

Finally, the Task-oriented Annotation Module $\mathcal{F}_a$ in Fig.\ref{fig:fig_2} (c) further provides detailed annotations, enabling the synthesized data to be applied to downstream tasks:

\begin{equation}
    \mathcal{A} = \mathcal{F}_a(\mathcal{I},  \hat{L}, T).
\end{equation}

\subsection{Task-specific Layout Generation Module}
\label{sec:layout}

Different tasks impose varying demands on image layouts: Simple tasks may only require generating reasonable layouts among given objects, while complex tasks may require adding, deleting, or modifying elements in the initial layout. In complex situations, a randomly generated layout falls short. Therefore, As Fig.\ref{fig:fig_2} (a) shows, we harness the reasoning ability of LLM to understand diverse inputs and generate the Initial Layout $L$, which is then adjusted by the statistical distribution $E$ from the reference dataset, obtaining an optimized Final Layout $\hat{L}$.

\noindent \textbf{Layout Initialization}. We sample from the object categories $O$, the initial bbox $B$, and the layout rule $R$ provided by the user, 
require the LLM to infer a reasonable overall scene $S$ based on the objects and layout rules. Next, The LLM needs to identify which objects with the label $l$ and reference modality $r$ are ultimately included after modifications to the initial layout and objects based on the layout rules. Through the reasoning on the relative positions, sizes, and relationships between instances, LLM finally summarizes and generates the \textbf{Initial Layout} $L$:

\vspace{-3pt}

\begin{equation}
    L = [S, L_{inst}], L_{inst} = \{ l, d, b, r \},
\end{equation}
we construct the basic scene $S$ and the instance layout $L_{inst}$, which includes the label $l$, detailed description $d$, bbox coordinates $b$, and reference modality $r$.

\noindent \textbf{Layout Adjustment.} The Initial Layout generated by the LLM may exhibit \textbf{issues on position and size}. For instance, in the first example of Fig.\ref{fig:layout_compare} (b), the generated person's size is inaccurate, resulting in only a partial person generated. The third example in Fig.\ref{fig:layout_compare} (b) shows that the LLM-generated Layout may have overly structured arrangements without diversity.
So we further use the statistical distribution $E$ of the reference dataset (such as COCO) to adjust the Initial Layouts $L$, aligning it more closely with the target distribution. 
As shown in the right half of Fig.\ref{fig:fig_2} (a), we adjust the size and position of the Initial layout. To perform a \textbf{reasonable size adjustment}, we analyze the normal distribution of the instance's width and aspect ratio in the reference dataset, with the sampled empirical width and aspect ratio, we adjust the initial layout size of the instance to some extent. Next, we randomly \textbf{move the center of current instance a certain distance in 8 directions}, and determine the best position by \textbf{minimizing the overlap ratio}, generating a more diverse and reasonable layout $\hat{L}$:

\vspace{-10pt}
\begin{equation}
    \hat{L} = [S, \hat{L}_{inst}], \hat{L}_{inst} = \{ l, d, f(b, E), r, \},
\end{equation}
in which $f$ presents the adjustment algorithm. Through the layout adjustment, we are able to modify the LLM-generated layout within reasonable limits, making it more aligned with the reference dataset distribution, while reducing overlap and improving the quality of the generated image. \textbf{Please refer to supplementary materials for details}.

\subsection{Uni-Controlled Image Generation} 
\label{sec:generation}

Specific tasks may introduce various conditions in generation. For example, users may want the generated image to follow a specific layout, align with a particular domain or contain an object in the reference image. In this section, we introduce the Uni-Controlled Image Generation Module, which is capable of generating high-quality images under various conditions.

\noindent \textbf{Multi-Conditioned Generation.} 
We leverage the control capabilities of MIGC~\cite{migc} to generate images with the layout $\hat{L}$. MIGC performs attention operations on the instance text features $f_t$ and image features $f_q$ of individual instances at the CA layer and EA layer, achieving separate shading of instances. Following the design of MIGC++~\cite{migc++}, we transform the reference image into the text features with ELITE~\cite{elite} global mapper and inject image information into the generation step.

In synthetic data, the data domain significantly affects the model's performance. Training on synthetic data with domains and styles different from the test set can degrade model performance. We attempt to \textbf{incorporate style information into the image generation process} to align the synthetic data more closely with the test set's domain. Inspired by StyleID~\cite{styleid}, we obtain the latent representation $x_{s,t}$ of the style image at each time step $t$ through DDIM Inversion~\cite{ddim}, and used them as keys and values for self-attention in specific layers, yielding style residuals $\phi_s$ that are then fused with content residuals $\phi_c$:

\vspace{-10pt}
\begin{equation}
    \phi_c = Attn(Q_c, K_c, V_c), \phi_s = Attn(Q_c, K_s, V_s),
\end{equation}

\vspace{-10pt}
\begin{equation}
    \phi = \lambda \phi_s + (1 - \lambda) \phi_c.
\end{equation}

This process can narrow the distance between domains with minimal overhead, while for domains with significant differences, fine-tuning the model still remains necessary.

\noindent \textbf{Quality and Position based Filtering.} To improve the quality of the synthesized data, we attempt to filter out low-quality content. Specifically,
We first use GLIP~\cite{glip1,glip2,glip3} and Grounding-DINO~\cite{gdino} to detect objects in the image and match the detected objects with the coordinates in the layout based on IoU. If neither detector identifies a particular instance, the image is discarded; otherwise, it is retained as a candidate. After collecting K candidate images, we use the confidence of the detector as the position score $P_i$, with the scores from the image evaluator~\cite{pickscore,aesthestic,imagereward,hps} as the visual quality score $Q_i$, the index of most suitable image $i^*$ is selected based on the sum of these scores:

\begin{equation}
    i^* = \arg \max_{i \in \{1, 2, \dots, K\}} (Q_i + P_i).
\end{equation}

\begin{table*}[!htb]
\setlength\tabcolsep{4pt}
\setlength{\abovecaptionskip}{0.cm}
    \setlength{\belowcaptionskip}{0.cm}
  \centering

  \begin{tabular}{c| c  c c c c | c c c c c | c c c c c}
    \toprule
    \multirow{2}{*}{\textbf{Method/shot}} & \multicolumn{5}{c|}{\textbf{Novel Set 1}} & \multicolumn{5}{c|}{\textbf{Novel Set 2}} & \multicolumn{5}{c}{\textbf{Novel Set 3}} \\

     &  1 &  2 & 3 &  5 & 10 &  1 &  2 & 3 &  5 & 10 & 1 &  2 & 3 &  5 & 10  \\
    \midrule
    FPD & 45.8 & 52.8 & 61.0 & 66.6 & 68.2 & 26.4 & 42.2 & 47.0 & 49.6 & 53.5 & 47.1 & 52.8 & 56.8 & 60.6 & 60.9 \\
    \midrule
    FPD + Ours & \textbf{63.6} & \textbf{65.5} & \textbf{67.1} & \textbf{68.2} & \textbf{69.1} & \textbf{44.8} & \textbf{48.5} & \textbf{51.9} & \textbf{52.8} & \textbf{55.0} & \textbf{55.4} & \textbf{57.8} & \textbf{59.6} & \textbf{61.6} & \textbf{63.8}  \\

    \bottomrule
  \end{tabular}
  \caption{ Quantitative results in VOC dataset in Few-Shot Object Detection Task. We present Novel AP50 of the augmentation results for three groups of Novel Sets based on different novel categories. Shot numbers (1, 2, 3, 5, 10) represent different shot settings.}
  \label{tab:FSOD-VOC}
\end{table*}

\begin{table}[!htb]
\setlength\tabcolsep{4pt}
\setlength{\abovecaptionskip}{0.cm}
    \setlength{\belowcaptionskip}{0.cm}
  \centering

  \begin{tabular}{c| c  c c| c c c}
    \toprule
    \multirow{2}{*}{\textbf{Method}} & \multicolumn{3}{c|}{10 shot} & \multicolumn{3}{c}{30 shot}\\

     &  AP50 &  AP75 & AP &  AP50 & AP75 &  AP   \\
    \midrule
    FPD & 34.2 & 12.9 & 15.8 & 40.4 & 17.5 & 19.6  \\
    FPD + ACP & 35.7 & 15.9 & 17.6 & 39.1& \textbf{19.3} & 20.1  \\
    \midrule
    FPD + Ours & \textbf{39.6} & \textbf{16.5} & \textbf{18.8} & \textbf{42.7} & 19.1 & \textbf{21.0}  \\

    \bottomrule
  \end{tabular}
  \caption{Quantitative results in COCO dataset in Few-Shot Object Detection Task. We present the AP50 for novel categories.}
  \label{tab:FSOD-COCO}
  \vspace{-5pt}
\end{table}

\subsection{Application to downstream Tasks}

\label{sec:region}

Specific tasks may require additional annotations. For example, instance segmentation tasks require mask annotations, while region captioning tasks require textual annotations for the region. Therefore, in this section, we introduce the annotation tools integrated for downstream tasks and summarize the complete process of generating various types of data using our framework.

\noindent \textbf{Task-Oriented Annotation.} Since generative models cannot strictly generate instances that perfectly fit the box, we use G-DINO~\cite{gdino} to refine the object bounding box coordinates (We called \textbf{Post-refine}). Next, based on the optimized layout, we generate images and text annotations for the synthesized data. For example, We \textbf{set a several Image Annotators}, including SAM\cite{sam} to obtain the instance masks and models like x2-upscaler to generate high-resolution images. We also utilize several \textbf{Analyzer to generate text annotations}: We generate global captions of the images using LLAVA\cite{improvedllava} and BLIP\cite{blip}. Using egtr\cite{egtr}, we detect relationships between instances. DynRefer\cite{dynrefer} helps us obtain region captions and instance attributes. We then utilize the VQA model\cite{ovis} to filter and enrich the above annotations, which inputs the detected annotations into a fixed template for filtering in the form of VQA. Finally, through LLM\cite{qwen}, we can summarize specific details of the generated images, thus creating question-answer pairs or other specialized annotations needed for specific tasks.

\noindent \textbf{Summarization.} 
Overall, our framework possesses universal data synthesis capabilities at each step of the generation process. When generating layouts, users typically only need to specify the object categories to sample. The introduction of layout rules also makes it possible to adapt to more diverse task forms, allowing users to specify image layouts according to their needs. During image generation, the inclusion of reference images and style images further enhances the versatility of the module. When users want to generate images from a particular domain, the simplest approach is to provide a reference style path, which is then incorporated into the image generation process. For datasets with significant domain differences, it may be necessary to replace the base generative model weights. Finally, for special annotations, we have integrated a series of tools. With LLM, user can process all the special annotation generated into question-answer pairs and other high-level annotations.

%% file: sec/4_experiment.tex
\section{Experiments}
\label{sec:exp}
\newcommand{\imporve}{\textcolor{green}}
\newcommand{\down}{\textcolor{red}}


\begin{table*}[!htb]
\setlength\tabcolsep{4pt}
\setlength{\abovecaptionskip}{0.cm}
    \setlength{\belowcaptionskip}{0.cm}
  \centering

  \begin{tabular}{c c | c  c c c | c c c c| c c c}
    \toprule
    \multicolumn{2}{c|}{\textbf{Benchmark}} & \multicolumn{4}{c|}{\textbf{CIRCO}} & \multicolumn{7}{c}{\textbf{CIRR}} \\

    \midrule
    
    \multicolumn{2}{c|}{\textbf{Metric}} & \multicolumn{4}{c|}{\textbf{mAP@K}} & \multicolumn{4}{c|}{\textbf{Recall@K}} & \multicolumn{3}{c}{\textbf{Recall$_{Subset}$@K}}  \\
    \midrule
    
     Backbone &  Method &  k=5 & k=10 &  k=25 & k=50 & k=1 & k=5 &  k=10 & k=50 & k=1 & k=2 & k=3  \\
    \midrule
    \multirow{5}{*}{ViT-L/14} & SEARLE & 11.68 & 12.73 & 14.33 & 15.12 & 24.24 & 52.48 & 66.29 & 88.84 & 53.76 & 75.01 & 88.19 \\
     & SEARLE-OTI & 10.18 & 11.03 & 12.72 & 13.67 & 24.87 & 52.31 & 66.29 & 88.58 & 53.80 & 74.31 & 86.94  \\
     & CIReVL & 18.57 & 19.01 & 20.89 & 21.80 & 24.55 & 52.31 & 64.92 & 86.34 & 59.54 & 79.88 & 89.69  \\
     &  LDRE & 23.35 & 24.03 & 26.44 & 27.50 & 26.53 & 55.57 & 67.54 & 88.50 & 60.43 & 80.31 & 89.90  \\
     &  \textbf{LDRE + Ours} & \textbf{24.57} & \textbf{25.65} & \textbf{28.08} & \textbf{29.17} & \textbf{27.33} & \textbf{55.74} & \textbf{68.15} & \textbf{88.92} & \textbf{62.51} & \textbf{81.49}  & \textbf{90.10}  \\

    \bottomrule
  \end{tabular}
  \caption{ Quantitative results in Zero-Shot Composed Image Retrieval Task.}
  \label{tab:ZSCIR}
\end{table*}

\begin{table}[!htb]
\setlength\tabcolsep{3pt}
\setlength{\abovecaptionskip}{0.cm}
    \setlength{\belowcaptionskip}{0.cm}
  \centering

  \begin{tabular}{c| c c  | c c | c c}
    \toprule
    \multirow{2}{*}{\textbf{Method}} & \multicolumn{2}{c|}{\textbf{animal}} & \multicolumn{2}{c|}{\textbf{aquarium}} & \multicolumn{2}{c}{\textbf{people-in-art}} \\

     &  AP &  AP50 & AP &  AP50 & AP &  AP50\\
    \midrule
    yolov5 & 84.7 & 57.2 & 80.9 & 47.4 & 52.7 & 21.2 \\

    yolov5 + Ours & \textbf{91.7} & \textbf{66.6} & \textbf{81.5} & \textbf{49.9} & \textbf{54.6} & \textbf{22.7} \\
    \midrule
    yolov7 & 51.1 & 30.9 & 81.2 & 51.1 & 67.3 & 27.2 \\
    yolov7 + Ours & \textbf{92.2} & \textbf{70.4} & \textbf{82.1} & \textbf{51.9} & \textbf{67.8} & \textbf{27.7}\\

    \bottomrule
  \end{tabular}
  \caption{ Quantitative results in Cross-Domain Object Detection. We selected the animal, aquarium, and people-in-art datasets. }
  \vspace{-5pt}
  \label{tab:CDOD}
\end{table}

We validate the effectiveness of our \ModuleName\ in four different scenarios.  In Section \ref{sec:setting}, we introduce the tasks, datasets, experimental setup, and some experimental details involved in our experiments. In section \ref{result}, we analyze the experimental results and conclusions across different tasks to demonstrate the effectiveness of our framework. In Section \ref{ablation}, we conduct ablation studies to verify the role of each module in our framework.

\subsection{Experimental Setup}
\label{sec:setting}

In this section, we introduce our experimental setup, the datasets, baselines and specific implementation details.

\noindent \textbf{Task Settings.} We validate the effectiveness of our framework across multiple tasks in \textbf{four different settings}:

\begin{itemize}
  \item [\textbf{$\bullet$}]
  \noindent \textbf{Few-shot Recognition.} We first evaluate the framework's ability to address \textbf{data scarcity} in \textbf{Few-shot Recognition} through two tasks: \textbf{1) Few-shot Object Detection.} Using the VOC~\cite{voc} and COCO~\cite{coco} datasets, we follow the default configuration, splitting the categories into base and novel with K images per category for training. For each novel category, we generate an additional 20 augmented images with the FPD~\cite{fpd} method as the baseline. \textbf{2) Few-shot Image Classification.} Following DA-Fusion~\cite{dafusion} and DIAGEN~\cite{diagen}, we select FOCUS~\cite{focus} and Custom-COCO~\cite{diagen} dataset, each category image has only K training images, we train the model in DIAGEN with 10 additional generated images for each category. 
  
  \item [\textbf{$\bullet$}]
  \noindent \textbf{Zero-Shot Composed Image Retrieval.} In ZSCIR task, users retrieve target images that match relevant images and relative captions. We believe constructing images to enhance retrieval accuracy will showcase our method's adaptability to complex scenarios. We used LDRE~\cite{ldre} with CLIP-L/14~\cite{CLIP} as the baseline model and tested on the CIRCO~\cite{circo} and CIRR~\cite{cirr} datasets, generating target images aligned with the query image content and relevant captions. The final retrieval similarity is computed as the weighted sum of the baseline and the similarity of generated with retrieval set images.

  \item [\textbf{$\bullet$}]
  \noindent \textbf{Cross-domain Detection.} The data domain plays a crucial role in model training and performance, so we selected several cross-domain datasets to validate the performance of our framework: \textbf{1) Cross-Domain Object Detection.} We first focused on the Roboflow dataset~\cite{roboflow}, which includes various domain-specific object detection datasets, and specifically targeted three: animal, aquatics, and people-in-art, representing real-world, underwater, and art domains. For each domain, we generated 1k training images and then trained YOLOv5~\cite{yolov5} and YOLOv7~\cite{yolov7} models. \textbf{2) Few-Shot Cross-Domain Object Detection.} We selected clipart dataset~\cite{clipart} on Few-Shot Cross Domain Object Detection~\cite{cdfsod}. Clipart is a cartoon-style dataset that annotates 20 categories corresponding to VOC. For each category, we generated 20 images. We use CD-ViTO~\cite{cdvito} as baseline.
    
  \item [\textbf{$\bullet$}]
  \noindent \textbf{Multi-modal Understanding.} We also test the effectiveness of our framework's synthesized data on the following tasks with multi-modal large model: \textbf{1) Multi-modal Image Perception.} We selected the LLaVA-7B model\cite{improvedllava} as our baseline, generated 5K training images, and fine-tuned directly on the pre-trained LLaVA-7B model. We tested model performance on the MME benchmark\cite{mme}. \textbf{2) Region Captioning.} We chose the GaLMM model \cite{glamm} as the baseline, generated 1K images with dynrefer-generated region captions, and combined them with the original training set, training the model under the default settings like learning rate and epochs. \textbf{3) Referring Expression Segmentation.} RES task requires the model to segment instances based on specific descriptions. We use the same 1k data in the RC task with SAM-generated mask to fine-tune directly on GaLMM trained in the first stage. For instance descriptions, we use the label (+label) or dynrefer-generated captions. 
\end{itemize}

\noindent \textbf{Implement Details.} For all experiments, we use their default configurations and parameters for training and inference. The base generator we use is MIGC, with RV as the foundation weights. In our experiments, we use Qwen1.5-32B as the LLM. All our experiments are conducted using PyTorch on four A6000 GPUs.

\subsection{Experiments Results}
\label{result}

\textbf{Few-shot Object Detection.}
Tab.\ref{tab:FSOD-VOC} and Tab.\ref{tab:FSOD-COCO} show the result for novel categories on VOC and COCO datasets. The data synthesized by \ModuleName\ leads to significant improvements across all these datasets,
The results demonstrate the effectiveness of our framework in the fundamental tasks.

\begin{figure}[tb]
    \setlength{\abovecaptionskip}{-0.cm}
    \setlength{\belowcaptionskip}{-0.cm}
    \begin{center}
        \includegraphics[width=0.47\textwidth]{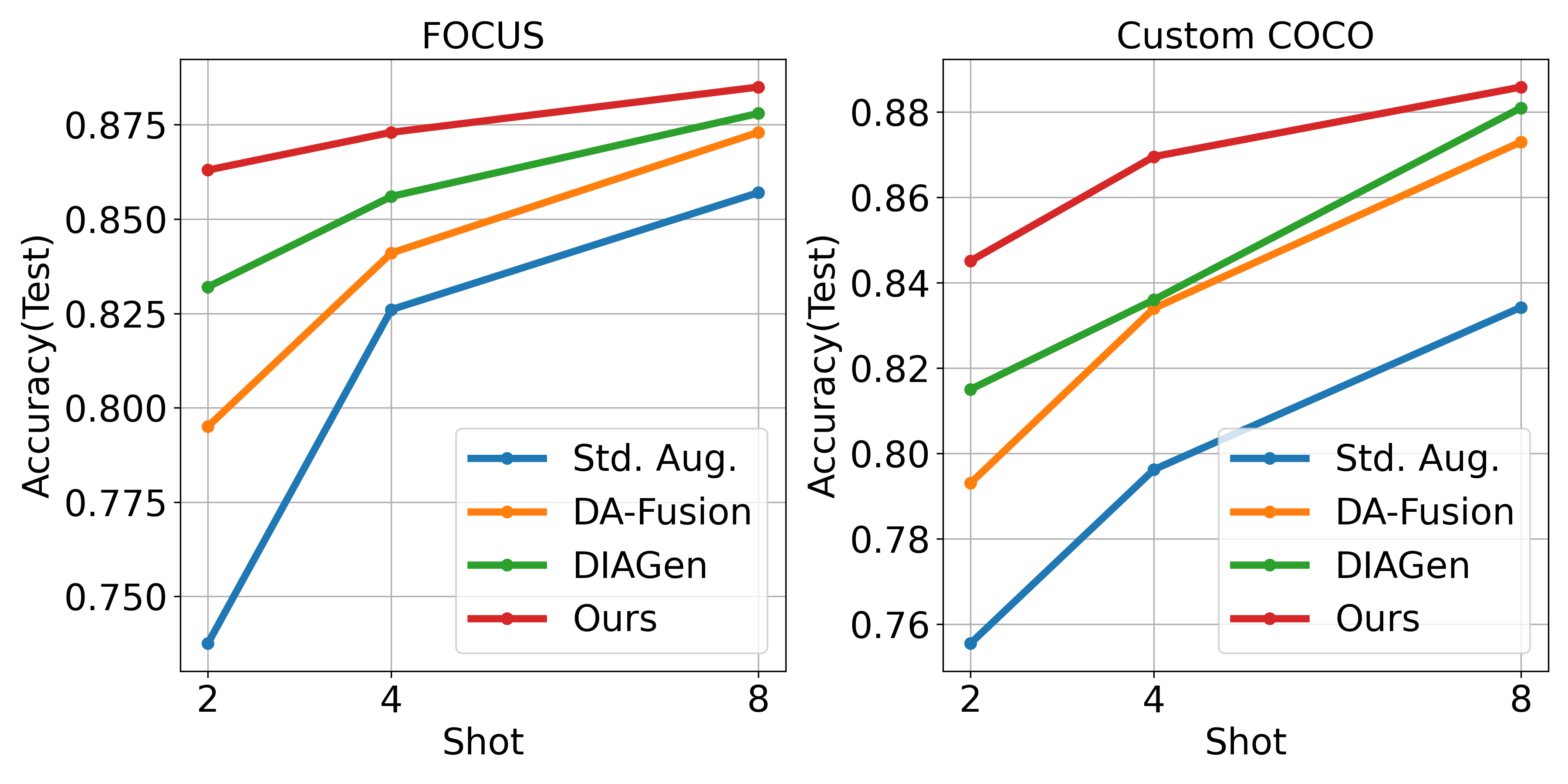}
        
    \end{center}

    \caption{ Quantitative results in Few-Shot Image Classification. } 
    
    \label{fig:fsclassification}
    \vspace{-10pt}
  \end{figure}

\noindent
\textbf{Few-shot Image Classification.} 
As shown in Fig.\ref{fig:fsclassification}, with the additional images generated by our framework, the performance is significantly improved. Besides, our method outperforms the DIAGEN, which also generates additional images, demonstrating that \ModuleName\ is more effective.

\begin{figure}[tb]
    \setlength{\abovecaptionskip}{-0.cm}
    \setlength{\belowcaptionskip}{-0.cm}
    \begin{center}
        \includegraphics[width=0.48\textwidth]{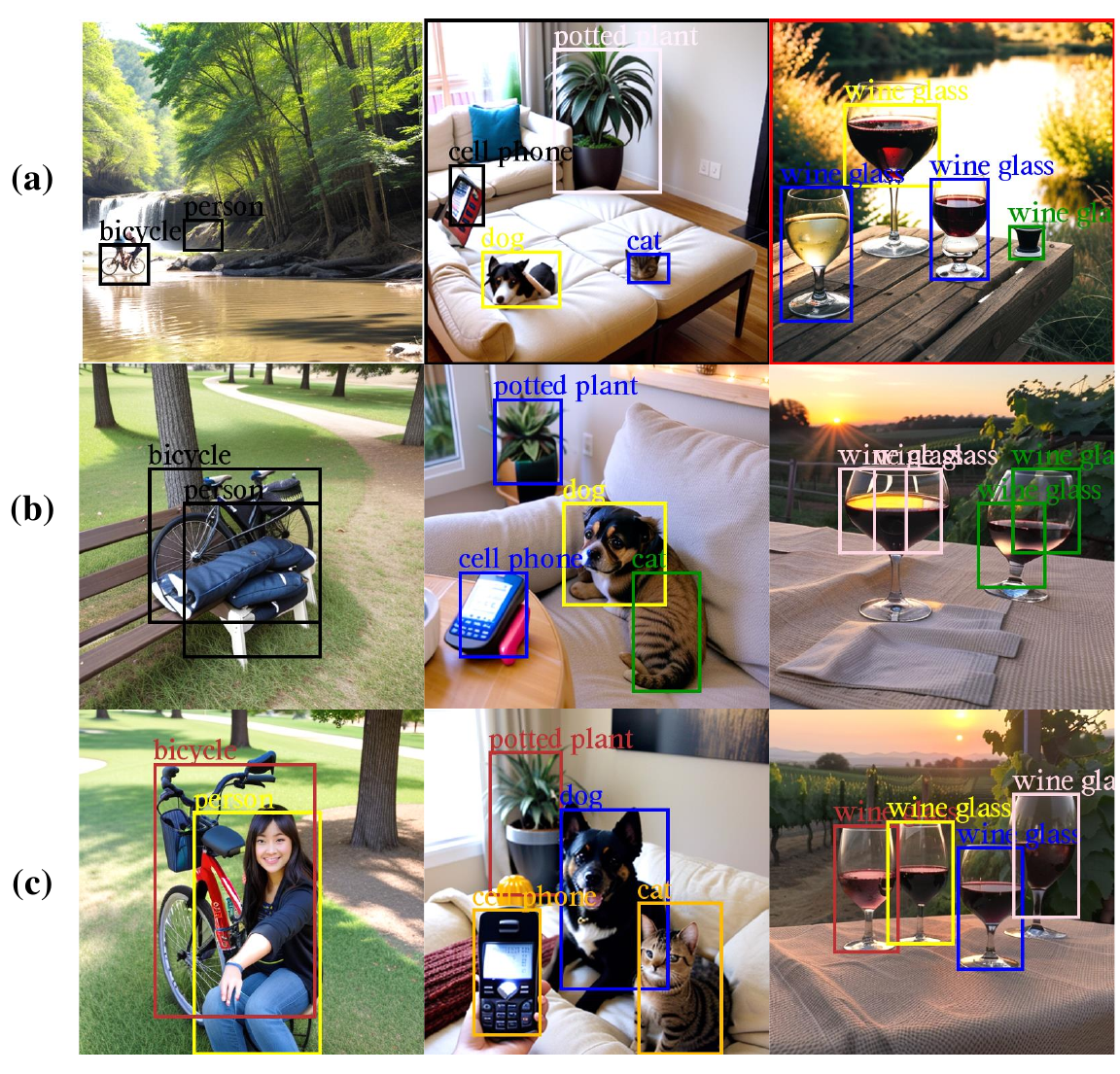}
        
    \end{center}
    \caption{ \textbf{Layout visualization results.} (a) corresponds to a randomly generated layout, (b) corresponds to the Initial layout generated by LLM, and (c) corresponds to the layout after adjustment. } 
    \label{fig:layout_compare}
    \vspace{-5pt}
  \end{figure}

\noindent \textbf{Zero-Shot Composed Image Retrieval.}  As shown in tab.\ref{tab:ZSCIR}, images constructed using our framework assist in retrieval, improving retrieval accuracy on both the CIRR and CIRCO datasets. This indicates that the images we generate effectively combine the original image content with relevant text information, providing more useful data for retrieval.

\noindent \textbf{Cross-Domain Object Detection.} The results in Tab.\ref{tab:CDOD} demonstrate that our method can adapt to tasks with domain differences and achieve performance improvements.

\begin{figure}[htbp]
    \begin{minipage}[c]{0.22\textwidth} 
        \centering
        \includegraphics[width=\textwidth]{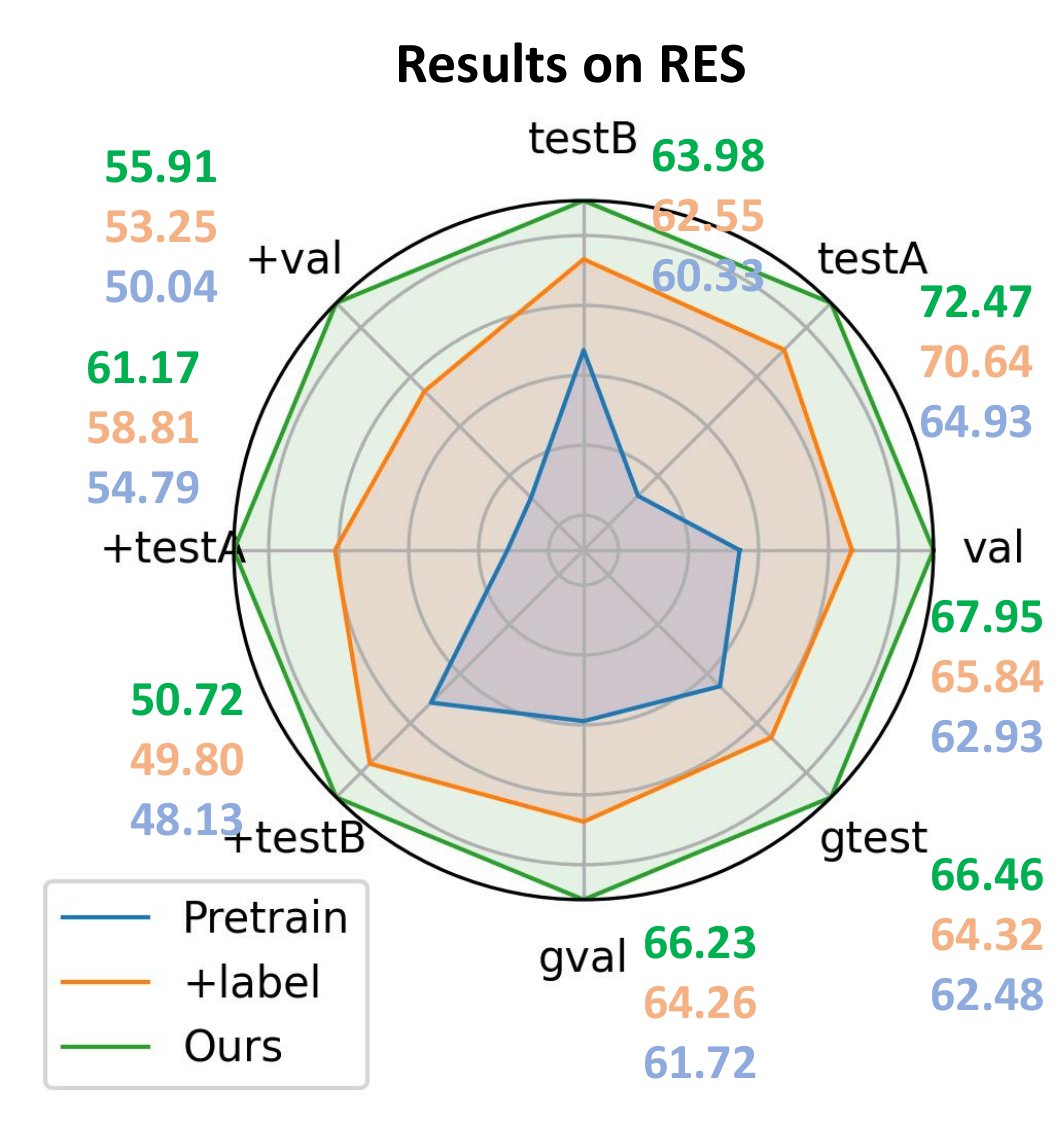}  
        \caption{Results in RES task, +label represents only use label as description.}
        \label{fig:RES}
    \end{minipage}
    \hspace{0.01\textwidth}  
    \begin{minipage}[c]{0.22\textwidth}  
        \centering
        \includegraphics[width=\textwidth]{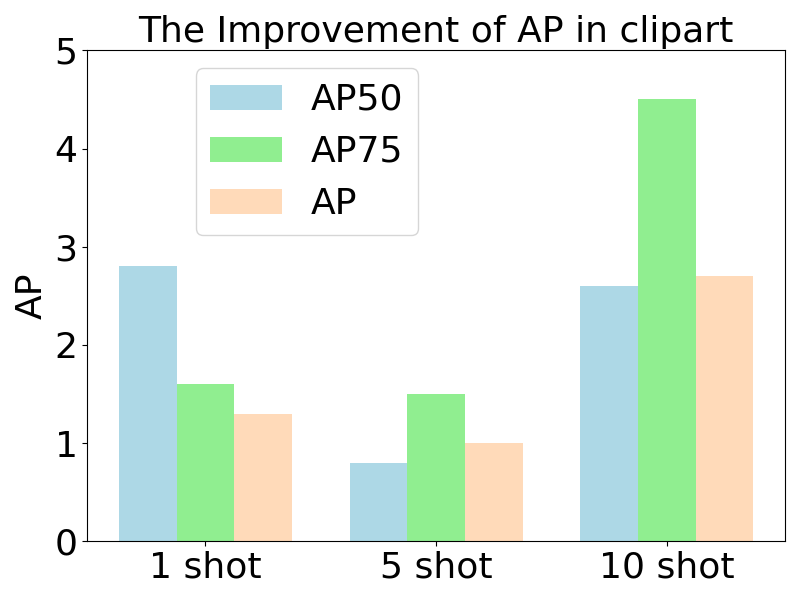}  
        \caption{Improvement in clipart in Few-shot Cross-Domain detection task.}
        \label{fig:clipart}
    \end{minipage}
\end{figure}

\begin{figure}[tb]
    \setlength{\abovecaptionskip}{-0.cm}
    \setlength{\belowcaptionskip}{-0.cm}
    \begin{center}
        \includegraphics[width=0.48\textwidth]{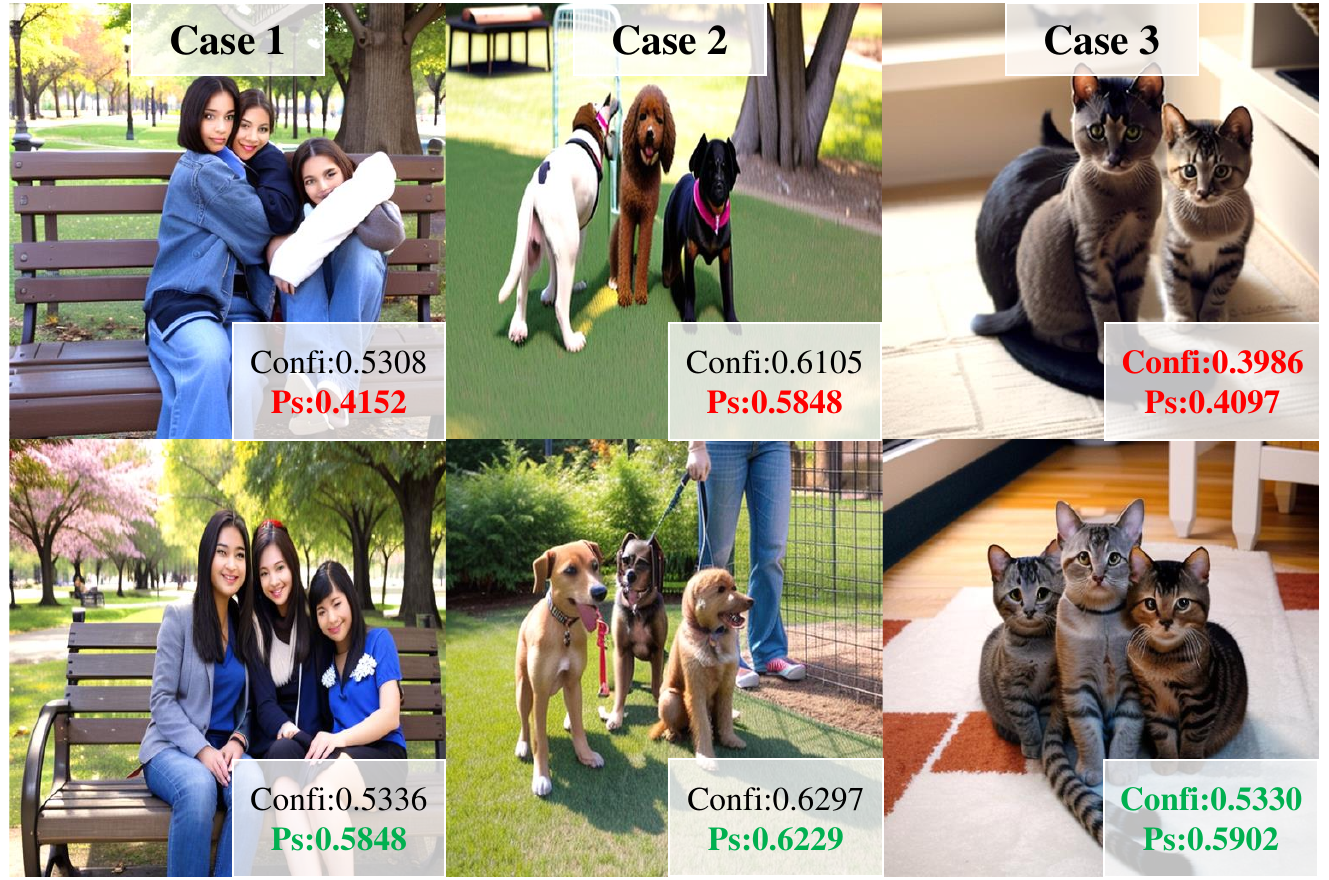}
        
    \end{center}
    \caption{ \textbf{Qualitative results}. Based on Position and quality scores, we can filter the low-quality and low-discriminability images. The Confi indicating the confidence of the detector, and Ps indicating the score of Pick-a-Score model. } 
    \label{fig:filter}
    \vspace{-10pt}
  \end{figure}

\begin{table}[!htb]
\setlength\tabcolsep{1pt}
\setlength{\abovecaptionskip}{0.cm}
    \setlength{\belowcaptionskip}{0.cm}
  \centering

  \begin{tabular}{c c c | c  c c| c c c}
    \toprule
    \multicolumn{3}{c|}{\textbf{Method}} & \multicolumn{3}{c|}{10 shot} & \multicolumn{3}{c}{30 shot}\\

    Layout & Filter & base &  AP50 &  AP75 & AP &  AP50 & AP75 &  AP   \\
    \midrule
     \multicolumn{3}{c|}{FPD} & 34.2 & 12.9 & 15.8 & 40.4 & 17.5 & 19.6  \\
    \midrule
    rand. & w/o & MIGC & 36.6 & 15.6 & 17.5 & 41.0 & 18.0 & 20.2  \\
    llm$^{\dagger}$ & w/o  & MIGC & 34.9 & 13.3 & 16.0 & 39.7 & 15.2 & 18.5  \\
    llm & w/o  & MIGC & \textbf{39.8} & 15.6 & 18.5 & \textbf{42.9} & 18.3 & \underline{20.8}  \\
    \midrule
    llm & quality & MIGC & 39.2 & 15.7 & 18.5 & 42.2 & 17.9 & 20.5  \\
    llm & position & MIGC & \textbf{39.8} & 15.6 & 18.7 & 42.3 & \underline{18.9} & 20.7  \\
    
    \midrule
    llm & both & GLIGEN & 38.1 & 15.9 & 18.2 & 41.5 & 18.8 & 20.6  \\
    llm & both & InstDiff & 38.9 & \textbf{17.0} & \textbf{19.1} & 41.0 & \underline{18.9} & 20.5  \\
    \midrule
    llm & both & MIGC & \underline{39.6} & \underline{16.5} & \underline{18.8} & \underline{42.7} & \textbf{19.1} & \textbf{21.0}  \\

    \bottomrule
  \end{tabular}
  \caption{Ablation result on COCO Few-shot Object Detection. llm$^{\dagger}$ represents using llm layout without Post-refine.}
  \label{tab:Ablation}
\end{table}

\noindent \textbf{Few-Shot Cross-Domain Object Detection.} The improvement of results are shown in Fig.\ref{fig:clipart}, which indicates that our framework is capable of generating images in the cartoon-style domain to improve the detection performance.

\begin{table}[htbp]
\setlength\tabcolsep{1pt}
    \begin{minipage}[c]{0.22\textwidth} 
        \centering
        \begin{tabular}{c| c c c }
            \hline
            {\textbf{Method}} & METEOR  & CIDEr \\
            \hline
            GaLMM & 16.1  & 106.0 \\
            +Ours & \textbf{16.7} & \textbf{107.6} \\
            \hline
        \end{tabular}
        \caption{Result in Region Captioning.}
        \label{tab:RC}
    \end{minipage}
    \hspace{0.01\textwidth}  
    \begin{minipage}[c]{0.22\textwidth}  
        \centering
        \begin{tabular}{c|c}
            \hline
            Method & MME \\
            \hline
            LLaVA-7B & 1507.9  \\
            LLaVA-7B + Ours & \textbf{1530.1}  \\
            \hline
        \end{tabular}
        \caption{Result of LLAVA on MME benchmark}
        \label{tab:LLAVA}
    \end{minipage}
    \vspace{-5pt}
\end{table}

\noindent \textbf{Multi-modal Large model.} The result in Tab.\ref{tab:LLAVA} shows that the data generated by our framework improves the model's performance on MME benchmark. In Tab.\ref{tab:RC}, we improve the performance of GLaMM on METOR from 16.1 to 16.7, and CIDEr from 106.0 to 107.6. Fig.\ref{fig:RES} shows that providing more detailed descriptions through the Task-Oriented Annotation module leads to greater improvements.

\subsection{Ablation Study}
\label{ablation}

\noindent
\textbf{The Impact of Image Layout on Performance.} We compare layouts generated by the Task-Specific Layout Module with those produced randomly on the COCO dataset for few-shot object detection. As shown in Tab.\ref{tab:Ablation}, randomly generated layouts provide a moderate performance improvement, with LLM generated layout, the improvement is further enhanced. Besides, we also visually present the images generated by different layouts. Fig.\ref{fig:layout_compare} (a) shows the images generated based on random layouts, which have issues with both position and size, with lower image quality. In Fig.\ref{fig:layout_compare} (b), we present the layouts generated by LLM. Although the position and relative sizes are more reasonable, there are still problems in size of objects like humans, and the position may be arranged too uniformly. In Fig.\ref{fig:layout_compare} (c), we show the images generated based on adjusted Final Layout, which shows the highest visual quality.

\noindent
\textbf{The Impact of Quality and Position-based Filtering.} We tested the impact of quality and position-based filtering in the COCO few-shot object detection. We generated K=4 candidate images and applied four filtering criteria: no filtering, position-bassed, quality-based, and comprehensive filtering. The results, as shown in Tab.\ref{tab:Ablation}, indicate that using comprehensive filtering yields the best AP results. In Fig.\ref{fig:filter}, we are able to \textbf{filter out the discordant or low-quality images} from case 1 and case 2 through quality scoring. In case 3, the confidence-based position scoring can filter out images with \textbf{low discernibility}.

 \noindent
\textbf{The Impact of Post-Refine.} We tested the impact of Post-refine in the COCO few-shot object detection. The llm$^{\dagger}$ in Tab.\ref{tab:Ablation} shows the result of training on data generated with LLM-generated layout but without post-refine, with performance decreased significantly.

\begin{figure}[tb]

    \setlength{\abovecaptionskip}{-0.cm}
    \setlength{\belowcaptionskip}{-0.cm}
    \begin{center}
        \includegraphics[width=0.48\textwidth]{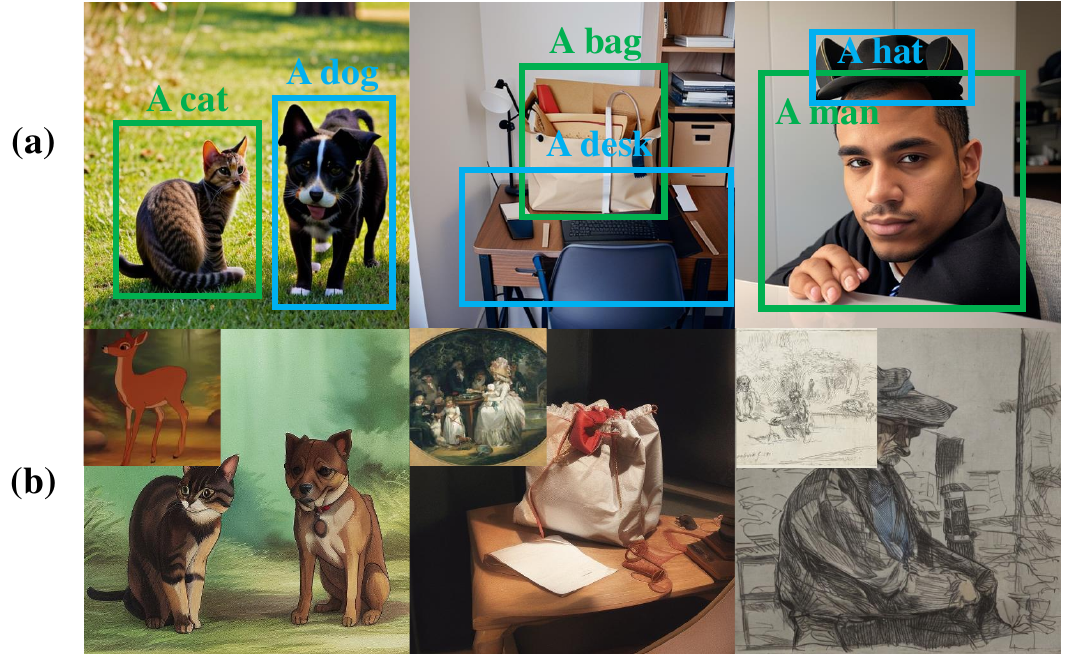}
        
    \end{center}

    \caption{ We present a comparison between (a) the image generated without style injection, and (b) the image generated with style injection. The image on the top-left of (b) is style image.} 

\vspace{-10pt}
    \label{fig:Style}
  \end{figure}

\noindent
\textbf{The Impact of Different Generators on Performance.} We generated training data with GLIGEN, InstanceDiffusion and MIGC on the Few-shot COCO Detection. The last three rows in Tab.\ref{tab:Ablation} show that all methods improve model performance, while InstanceDiffusion and MIGC, with stronger control abilities, bring more improvements.

\noindent
\textbf{Compare with other Framework.} We have already compared our Framework with DIAGEN in the Few-shot image classification task and achieved stronger performance. In the COCO Few-shot object Detection, we compare our approach with the ACP framework. The result in Tab. \ref{tab:FSOD-COCO} shows that our approach achieves better augmentation results. We believe this is due to the more reasonable layouts and higher-quality filtering and control. 

\begin{table}[!htb]
\setlength{\abovecaptionskip}{0.cm}
    \setlength{\belowcaptionskip}{0.cm}
  \centering

  \begin{tabular}{c| c c  }
    \toprule
    \textbf{Method} & AP50 & AP  \\

    \midrule
    yolov5 & 52.7 & 21.2 \\

    w/o align &  51.8 \textcolor{red}{(-0.9)} & 20.1 \textcolor{red}{(-1.1)} \\
    \midrule
    w/ align & \textbf{54.6} \textcolor{green}{(+1.9)} & \textbf{22.7} \textcolor{green}{(+1.5)}\\

    \bottomrule
  \end{tabular}
  \caption{ Ablation of domain align on the people-in-art dataset.}
  \vspace{-5pt}
  \label{tab:ablation_cd}
\end{table}

\noindent
\textbf{The Impact of Domain Alignment on Performance.} We evaluate domain effects on performance in the cross-domain people-in-art dataset. As shown in Tab.\ref{tab:ablation_cd}, unaligned domain generation affects training performance, while alignment leads to improvements. We also show the generated images with style control in Fig.\ref{fig:Style}, which indicates that our framework is able to control the position while aligning the domain of the generated image with the reference image.

%% file: sec/5_conclusion.tex
\section{Conclusion}
\label{sec:conclusion}

In this paper, we propose  \ModuleName, a universal data synthesis framework. We use three modules to progressively generate target data in the `Layout-Image-Annotation' format. We use task-specific layout generation modules to create layouts that meet task requirements, with the Uni-Controlled Image Generation module to generate high-quality images based on the layout, reference image, and style image. The Task-oriented Annotation module further generates additional annotations required by specific tasks. A series of experiments in few-shot object detection, zero-shot composed image retrieval, cross-domain object recognition,  as well as multi-modal image perception and grounding demonstrate the effectiveness and versatility of our data synthesis framework.

%% file: sec/X_suppl.tex
\clearpage

\setcounter{page}{1}
\setcounter{section}{0}
\setcounter{figure}{0}
\setcounter{section}{0}
\setcounter{equation}{0}
\gdef\thesection{\Alph{section}}
\maketitlesupplementary

\section{Framework Details}
\label{sec:rationale}

\subsection{Task-Specific Layout Generation}

\begin{figure*}[tb]

    \setlength{\abovecaptionskip}{0.cm}
    \setlength{\belowcaptionskip}{0.cm}
    \begin{center}
        \includegraphics[width=1.0\textwidth]{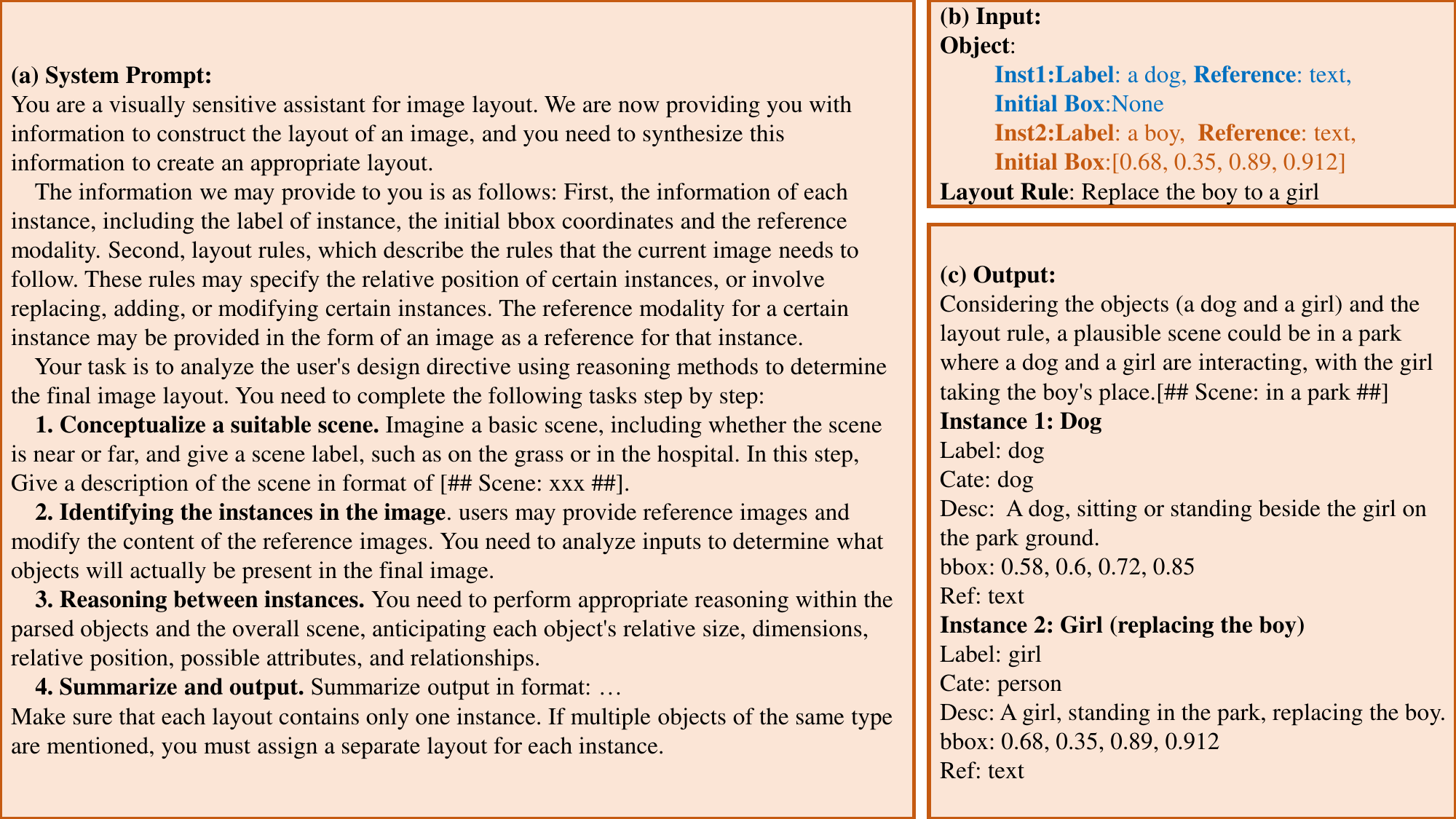}

    \end{center}
   \caption{\textbf{Overview of our system prompt, Input and output}.
} 
    \label{fig:layout_system}
  \end{figure*}

\noindent \textbf{Layout Initialization.} To achieve flexible and versatile layout generation, the Task-Specific Layout Generation module should be designed to accommodate a wide range of requirements. This includes the ability to infer suitable layouts based on the current objects and allows users to control objects' addition, deletion, modification, or arrangement through layout rules. Subsequently, the LLM~\cite{gpt4,qwen} generates the Initial Layout based on these rules to propose a baseline layout scheme for the objects. The system prompts we set during this process and the corresponding output format is shown in Fig.\ref{fig:layout_system}. Specifically, in addition to generating object labels and bbox coordinates, we generate the following additional attributes: \textbf{Cate}: A single word describing the general category of the object, used for subsequent position detection. \textbf{Desc}: A detailed category description, providing rich attributes on the textual side. \textbf{Ref}: Indicating whether to use reference image modality information in the subsequent generation process.

\noindent \textbf{Layout Adjustment.}
After generating the initial layout, we attempt to adjust it to make it more reasonable. We sample the width and aspect ratio distributions $E$ of each category from the target dataset or reference dataset (e.g., COCO~\cite{coco}) and adjust instance size and position: For each instance $i$ with an initial coordinate $b_i$, we sample the widths and aspect ratios 100 times from the distribution of its category $l_i$ and compute the average to obtain a robust empirical distribution $(w, a)$ that excludes outliers. We adjust the size of the instance by applying a weighted average (with weight sampled between 0.1 and 0.2 for empirical) of the empirical size and the LLM initialized size.
After obtaining the new instance bounding boxes, we attempt to adjust the position of each instance. Specifically, we slightly shift the center points of the object boxes in 8 different directions, then calculate the overlap ratio of the new bounding box within the revised layout. The direction with the smallest overlap ratio is selected for perturbation.

\subsection{Uni-Controlled Image Generation}

\noindent \textbf{Multi-conditioned Generation} Image generation under multi-conditional control is achieved by building on the MIGC framework~\cite{migc}. MIGC employs a divide-and-conquer strategy to render each instance individually, enabling precise control over positions and attributes. As a plug-and-play controller, MIGC is compatible with various generator weights and easily extendable. Based on this flexibility, we introduce additional control conditions to achieve image generation under multiple constraints. Specifically, following MIGC++~\cite{migc++}, we leverage ELITE\cite{elite} to map reference images into the textual feature space, incorporating visual modality information into the MIGC generation process. Inspired by StyleID~\cite{styleid}, we inject style information into the image generation pipeline. Concretely, we use DDIM inversion~\cite{ddim} to extract latent representations of the style image at each time step. During each step of MIGC’s generation process, the corresponding latent is used as the key and value to inject style information into specific layers of Self-Attention. The original generation output is then weighted and combined with the style-enhanced output. Considering that diffusion models focus more on layout in the early stages and details in the later stages, we apply style information with a higher weight ($\lambda$ = 0.7) during the early to mid stages (time steps 0-35) and a lower weight ($\lambda$ = 0.3) in the later stages.

\noindent \textbf{Quality and Position-Based Filtering} To ensure high-quality synthetic images, we observed that the framework occasionally generates low-quality images, reflected in overall image quality, the recognizability of individual instances, or the details of certain instances. To address this, we designed a specialized filtering scheme to enhance the quality of the constructed dataset. During the generation of each image, we use GLIP~\cite{glip1} and G-DINO~\cite{gdino} to detect instances and obtain their positions. These positions are matched with the input control layout. If the IoU between a detected instance and the corresponding layout instance exceeds 0.5, the instance is considered correctly generated; otherwise, it is marked as a failure. Images where all instances are correctly generated are retained as candidate images. Once K candidate images are obtained, we select the most suitable one based on quality and positional evaluation metrics. For instances that are difficult to recognize, we utilize the generator’s confidence scores. We calculate the average confidence scores from GLIP and G-DINO detections for the entire image and use this as the image’s confidence score. For overall or detailed image quality, we employ standard image quality assessors. The final selection is based on the sum of the quality and confidence scores, ensuring the highest-quality image is chosen.

\begin{figure*}[tb]

    \setlength{\abovecaptionskip}{0.cm}
    \setlength{\belowcaptionskip}{0.cm}
    \begin{center}
        \includegraphics[width=0.9\textwidth]{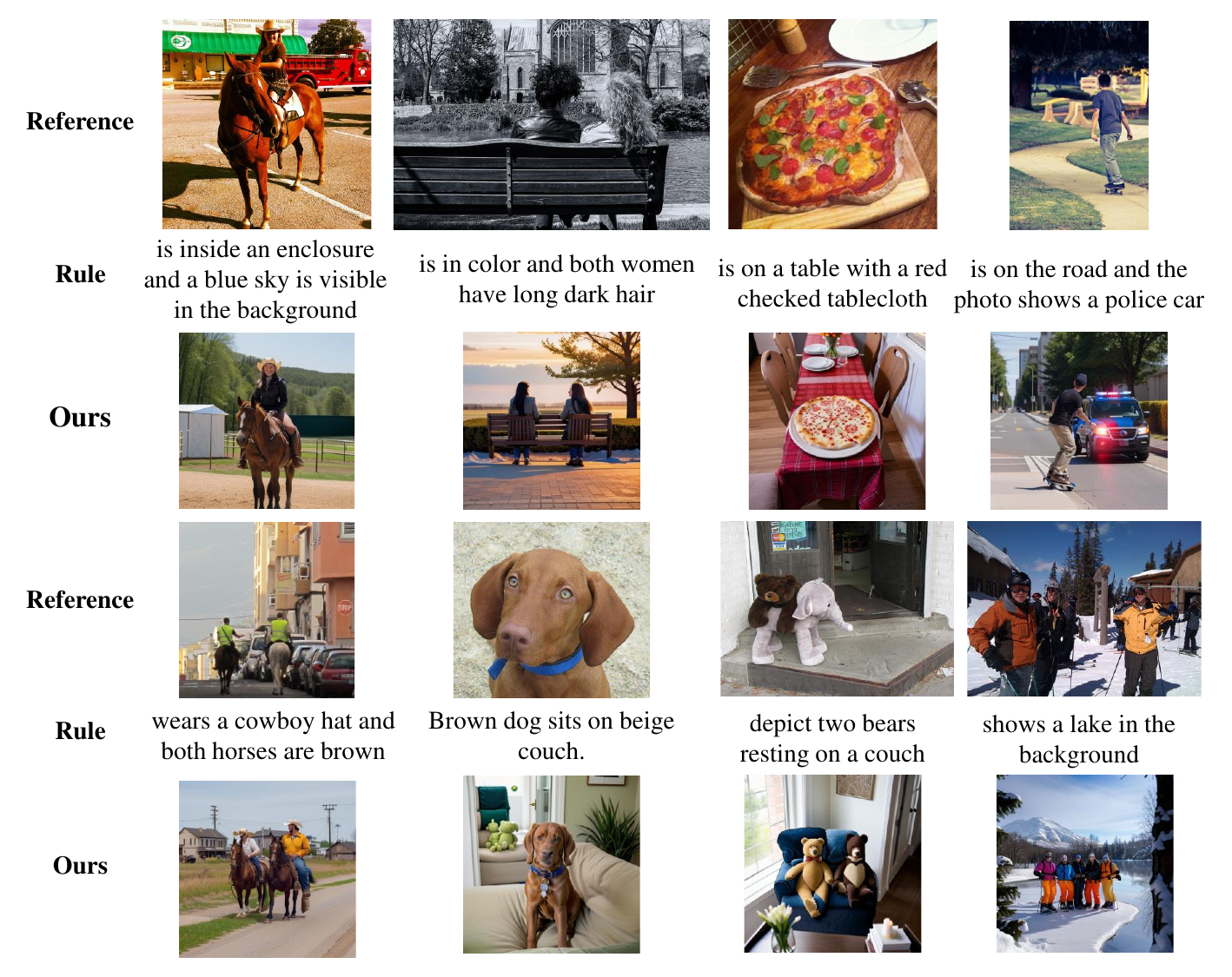}

    \end{center}
   \caption{\textbf{The Image generated by our Framework in ZSCIR.}
} 
    \label{fig:supp_zscir}
  \end{figure*}

\section{Experiments Details}

\noindent \textbf{Few-Shot Object Detection.}
The FSOD task divides the object categories in the dataset into Novel and Base classes. A base detector is trained on the Base classes and then fine-tuned on the Few-Shot dataset containing both Base and Novel classes with K images for each category. The performance on Novel classes is evaluated. We conducted experiments on the VOC~\cite{voc} and COCO datasets. VOC datasets divide the 20 classes into 15 base classes and 5 novel classes, with three different class partitions designed for comprehensive evaluation. The training process uses the train/val sets from VOC07 and VOC12, while the VOC07 test set is used for evaluation. Performance is measured using mAP50 under K-shot settings, where 
$K=\{1,2,3,5,10\}$. For MS COCO, the 20 PASCAL VOC classes are treated as Novel classes, while the remaining 60 classes are Base classes. The 5k images from the COCO 2017 validation set are used for evaluation, while the rest are used for training. Performance is reported using AP at IoU=0.5:0.95 under $K=\{10,30\}$-shot settings.

For both datasets, we select the FPD model~\cite{fpd} as the baseline, with ResNet-101 as the backbone. We directly use the pre-trained FPD detector on the Base classes, and fine-tune it on our augmented few-shot dataset. We generate 20 additional images for each Novel class, focusing on a single category per image, with each image containing 1–4 instances. All other experimental details follow the default configurations of the FPD framework, including epochs, learning rates, and training schedules.

\begin{figure*}[tb]
    \setlength{\abovecaptionskip}{0.cm}
    \setlength{\belowcaptionskip}{0.cm}
    \begin{center}
        \includegraphics[width=0.9\textwidth]{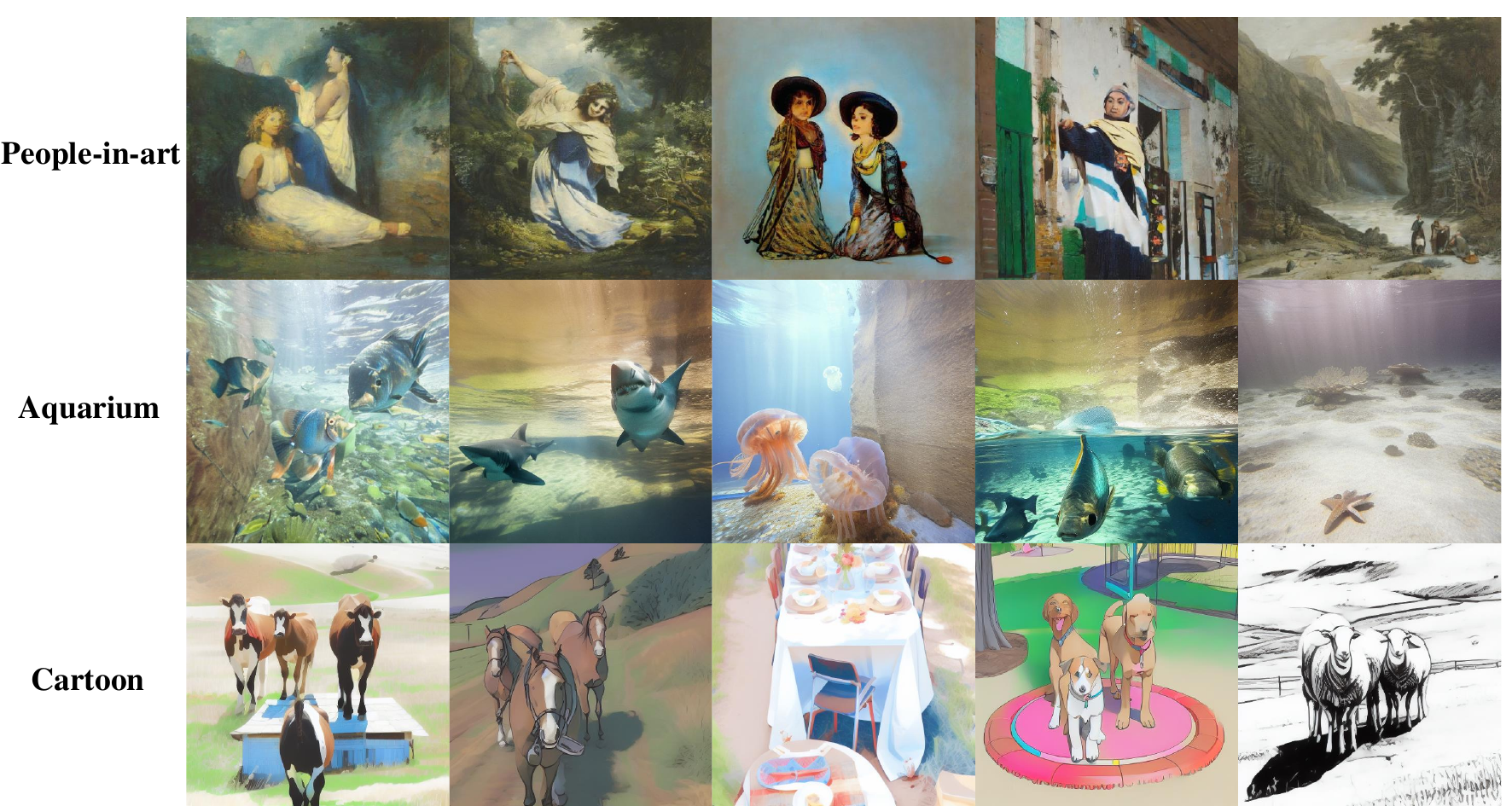}
    \end{center}
   \caption{\textbf{The Image generated by our Framework in Cross-Domain Detection.}
} 
    \label{fig:supp_cd}
  \end{figure*}

\noindent \textbf{Zero-Shot Composed Image Retrieval.}
We generate corresponding synthetic images for the test datasets of the ZSCIR tasks CIRCO~\cite{circo} and CIRR~\cite{cirr}. For each query, we sample a reference object image and the corresponding layout rule from the dataset. 
Using the mask generated by Grounded-SAM, we segment the main subject from the reference image, convert it into CLIP textual features using ELITE, and inject these features into the textual representation. We generate 5 synthetic images for each reference image as auxiliary retrieval images. During retrieval, we compute the similarity between each synthetic image and the retrieval dataset, averaging these similarities to obtain an auxiliary retrieval similarity for the query. This similarity is then combined with the baseline retrieval similarity using a weighted average.

\noindent \textbf{Cross-Domain Object Detection.}
For the dataset in different domains, we generate 1,000 synthetic images, with each image containing 1–4 objects from a specific instance category. For each detector, training is conducted for 100 epochs to ensure convergence, using default parameter settings, including learning rate and other configurations.

\noindent \textbf{Multi-Modal Image Perception.}
During training, we constructed 5,000 images, each containing 2–4 objects sampled from 2–3 COCO categories. We fine-tune the pre-trained LLAVA-7B model~\cite{llava} using the constructed dataset, setting the learning rate to $2 \times 10^{-5}$ and the projector learning rate to $1 \times 10^{-6}$, training for one epoch.

\noindent \textbf{Region Captioning.}
We construct 1,000 images, each containing 2–4 objects sampled from 2–3 COCO categories. Our synthetic data is combined with the original refCOCOg training data for training. Using a pre-trained one-stage model, we set the learning rate to $2 \times 10^{-5}$, train for 5 epochs, iterate 500 times per epoch, and use a batch size of 128. The synthetic and real data are sampled at a 3:1 ratio.

\noindent \textbf{Relation Extraction with Synthesis.}
For the RES task, we use the 1,000 images constructed for RC as training data. The synthetic data is directly used for training on a pre-trained one-stage model, following the same training configurations as RC. The performance improvements brought by training on the one-stage model demonstrate the effectiveness of our synthetic data.

\section{Visualization Results}
\label{sec:visual}

\subsection{Visualization Results in ZSCIR.}

We present the generated image in ZSCIR task in Fig.\ref{fig:supp_zscir}. In the figure, we present the results of generating target images using reference images and layout rules. It can be observed that the generated images retain some attribute information from the original images (e.g., the cowboy hat in the first row, first column, and the dog breed in the second row, second column). At the same time, our images also reflect the layout rules to a certain extent, including adding objects, modifying backgrounds (such as the sky, colorful backgrounds, or lake surfaces), and controlling the attributes of the main subjects.

\subsection{Visualization in Cross-domain Detection.}

We present the generated image in the Cross-domain Object Detection task in Fig.\ref{fig:supp_cd}. Specifically, we demonstrate the generation results in three rows, showcasing the effects in artistic style, underwater, and cartoon style, respectively.

\subsection{Visualization in Multi-modal Understanding.}

\begin{figure*}[tb]
    \setlength{\abovecaptionskip}{0.cm}
    \setlength{\belowcaptionskip}{0.cm}
    \begin{center}
        \includegraphics[width=1.0\textwidth]{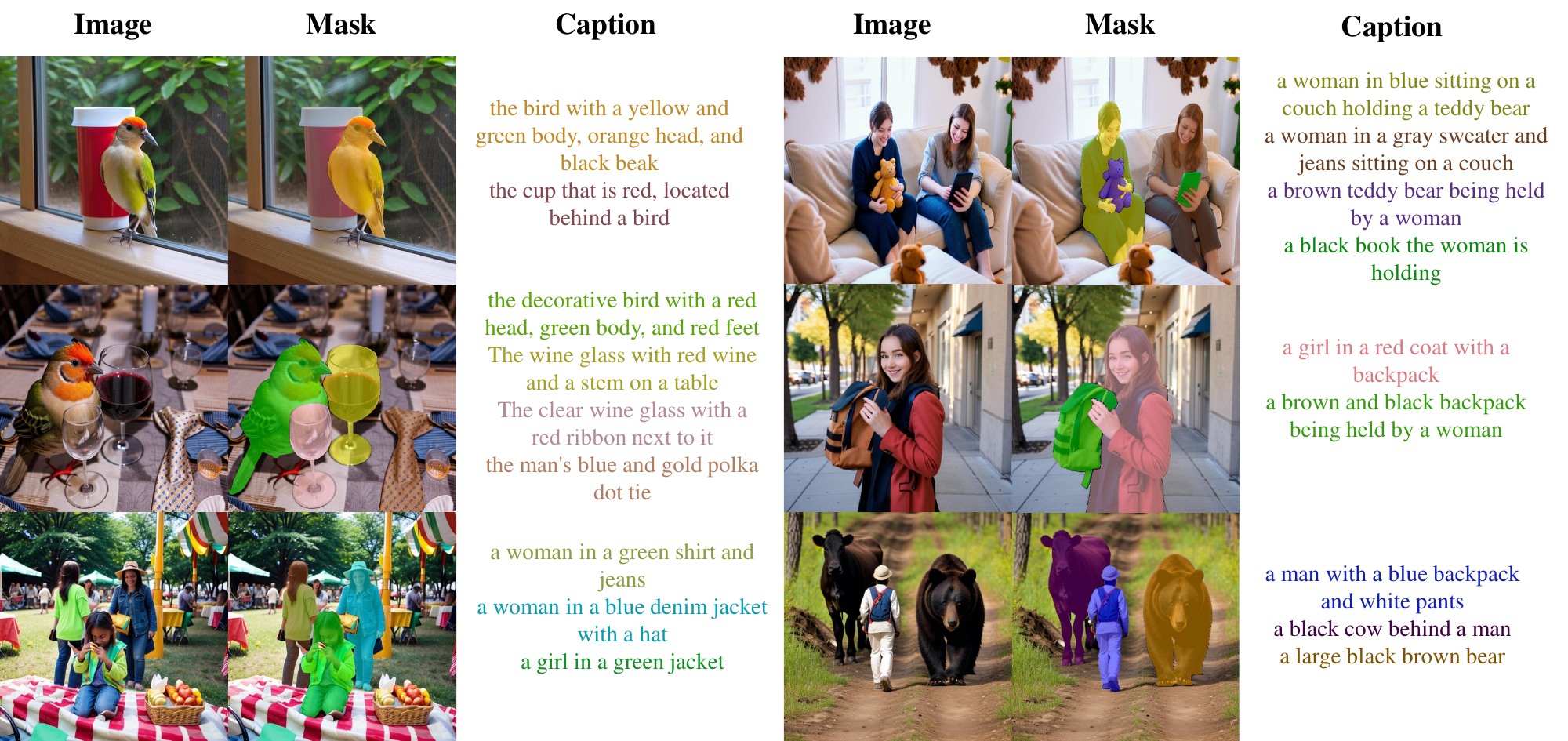}
    \end{center}
   \caption{\textbf{The Image generated by our Framework in Multi-modal Understanding.}
} 
    \label{fig:supp_mm}
  \end{figure*}

We present the generated image in the multi-modal Understanding Tasks in Fig.\ref{fig:supp_mm}, along with the corresponding annotations generated by the task-oriented annotation module. It can be observed that our generated images exhibit relatively reasonable layouts and, after filtering, achieve high quality. Additionally, the annotations constructed by the task-oriented module are also fairly reasonable. However, due to the limitations of the generator's performance, there are still some deficiencies in the details of the final regions in the generated images.

\section{Limitation and Future Work}
\noindent \textbf{Limitation.} Due to the generator's performance limitations, certain challenging categories still require fine-tuning the model on the target dataset to generate them effectively. 
For example, it is still challenging to generate realistic images in the field of medical imaging, as it requires corresponding data for fine-tuning.

\noindent \textbf{Future Work.}
In the future, we plan to explore applying our unified synthesis framework to tasks in specific domains with scarce data, such as medical imaging and remote sensing. We aim to generate data more closely aligned with the characteristics of each field. This approach aligns with our overarching goal of building a universal synthesis framework and represents a critical direction for future efforts.